\documentclass{article}
\usepackage{spconf,amsmath,graphicx}
\usepackage{xspace}
\usepackage{bm}
\usepackage{paralist}
\usepackage{amssymb}
\usepackage{booktabs}
\usepackage{multirow}
\usepackage{color}
\usepackage{amsfonts}
\usepackage{amsmath}
\usepackage[caption=false]{subfig}
\usepackage[dvipsnames]{xcolor}
\usepackage{float}
\usepackage[T1]{fontenc}
\usepackage{bbding}
\usepackage{cite}  

\ninept
\begin{document}
\title{KAnoCLIP: Zero-Shot Anomaly Detection through Knowledge-Driven Prompt Learning and Enhanced Cross-Modal Integration}

%
%

%
\name{Chengyuan Li$^{1, \dagger}$,  Suyang Zhou$^{1}$, Jieping Kong$^{1}$, Lei Qi$^{2}$, Hui Xue$^{2}$ \thanks{$^{\dagger}$ Corresponding author} }
  
\address{
    \textsuperscript{\textmd{1}} College of Software Engineering, Southeast University, Nanjing, China \\
    \textsuperscript{\textmd{2}} 
    School of Computer Science and Engineering, Southeast University, Nanjing, China\\
    $\left\{chengyuanli, suyangzhou, jiepingk, qilei, hxue\right\}@seu.edu.cn$
    }


%
\maketitle              
\begin{abstract}

     Zero-shot anomaly detection (ZSAD) identifies anomalies without needing training samples from the target dataset, essential for scenarios with privacy concerns or limited data. Vision-language models like CLIP show potential in ZSAD but have limitations: relying on manually crafted fixed textual descriptions or anomaly prompts is time-consuming and prone to semantic ambiguity, and CLIP struggles with pixel-level anomaly segmentation, focusing more on global semantics than local details. To address these limitations, We introduce KAnoCLIP, a novel ZSAD framework that leverages vision-language models. KAnoCLIP combines general knowledge from a Large Language Model (GPT-3.5) and fine-grained, image-specific knowledge from a Visual Question Answering system (Llama3) via Knowledge-Driven Prompt Learning (KnPL). KnPL uses a knowledge-driven (KD) loss function to create learnable anomaly prompts, removing the need for fixed text prompts and enhancing generalization. KAnoCLIP includes the CLIP visual encoder with V-V attention (CLIP-VV), Bi-Directional Cross-Attention for Multi-Level Cross-Modal Interaction (Bi-CMCI), and Conv-Adapter. These components preserve local visual semantics, improve local cross-modal fusion, and align global visual features with textual information, enhancing pixel-level anomaly detection. KAnoCLIP achieves state-of-the-art performance in ZSAD across 12 industrial and medical datasets, demonstrating superior generalization compared to existing methods.

\end{abstract}


%
%
%
\section{Introduction}

Anomaly detection (AD) ~\cite{roth2022towards,tian2023self} is crucial in fields like industrial quality assurance, medical diagnostics, and video analysis, encompassing anomaly classification (AC) and anomaly segmentation (AS) for image-level and pixel-level anomalies. The main challenges in AD are the rarity and diversity of anomalies, making dataset collection costly and time-consuming. Traditional one-class or unsupervised methods ~\cite{tian2023self} often fall short due to the diverse and long-tail distribution of anomalies across domains such as industrial defects and medical lesions, necessitating a generalizable model. Zero-shot anomaly detection (ZSAD) ~\cite{huang2024adapting,jeong2023winclip,zhou2023anomalyclip} is vital for identifying anomalies without extensive training samples, particularly when data privacy concerns or insufficient labeled data are issues. Vision-Language Models (VLMs), like CLIP \cite{radford2021learning}, have advanced ZSAD by leveraging training on large datasets of image-text pairs. 

However, CLIP's original model still faces challenges in anomaly detection due to the task's complexity. There are two main limitations in applying CLIP to zero-shot anomaly detection: (1) existing methods ~\cite{radford2021learning,jeong2023winclip,chen2023zero,huang2024adapting} rely on fixed text descriptions or anomaly prompts. For zero-shot anomaly detection using CLIP, a commonly used text prompt template is "a photo of a [class] with holes." However, these handcrafted prompts require extensive expertise, are time-consuming, and suffer from semantic ambiguity. Inspired by prompt learning in NLP, CoOp \cite{zhou2022learning} utilizes learnable vectors for prompts and requires only a few labeled images. Despite this, CoOp tends to overfit to base classes, thus diminishing its ability to generalize to unseen classes. 
(2) while CLIP aligns image-level semantics with anomaly prompts through cross-modal contrastive training, it performs poorly in precise anomaly segmentation (pixel-level detection) because it emphasizes global semantics and overlooks local details \cite{radford2021learning,huang2024adapting}. A mechanism for refining CLIP models in the local visual space and integrating local pixel-level cross-modal features is necessary for superior anomaly segmentation.

To address the two main limitations, we propose the KAnoCLIP framework, which introduces KnPL. This approach leverages large language model (LLM) and a visual question answering (VQA) system to generate anomaly descriptions, forming a knowledge base that guides the development of Learnable Normal Prompts (LNPs) and Learnable Abnormal Prompts (LAPs) using KD loss. This Loss minimizes the Euclidean distance between LLM-VQA-generated abnormal prompts and LAPs while maximizing the distance to LNPs, effectively eliminating the reliance on fixed text prompts and enhancing generalization to new anomaly classes. Furthermore, the framework integrates CLIP-VV, Bi-CMCI, and Conv-Adapter to preserve local visual semantics, improve cross-modal fusion, and align global visual features with textual information. These enhancements collectively improve the detection of subtle anomalies. Our main contributions are summarized as follows:
\begin{itemize}
    \item KAnoCLIP is a novel zero-shot anomaly detection solution that doesn't require training samples from the target dataset, making it ideal for applications with privacy concerns or limited data, especially in industrial and medical fields.
    \item We introduce knowledge-driven prompt learning to eliminate manual text prompting and alleviate overfitting, thereby enhancing generalization to new anomaly classes.
    \item KAnoCLIP integrates CLIP-VV, Bi-CMCI, and Con-Adapter to refine local visual spaces and enhance cross-modal interactions, achieving robust pixel-level anomaly segmentation.
    \item Extensive experiments on 12 industrial and medical datasets demonstrate that KAnoCLIP consistently outperforms state-of-the-art techniques, setting a new benchmark for ZSAD.
\end{itemize} 


 \begin{figure}[t!]
    \centering
    \includegraphics[width=\linewidth]{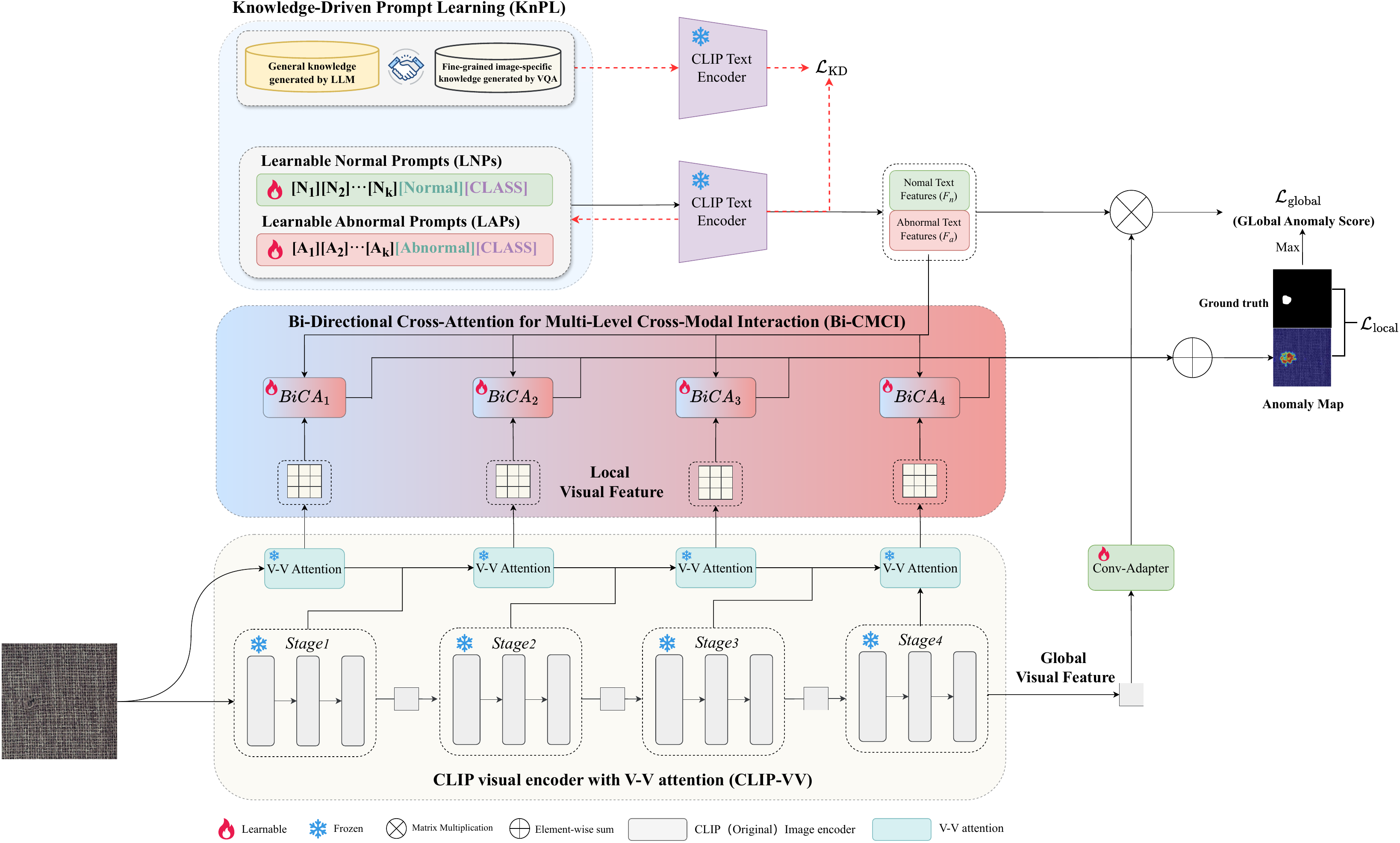}
    \caption{Illustration of KAnoCLIP, which consists of four key components: KnPL, CLIP-VV, Bi-CMCI, and Conv-Adapter. KnPL uses an LLM and VQA system to form an LLM-VQA knowledge base, guiding the generation of learnable normal and abnormal prompts (LNPs and LAPs), reducing overfitting and enhancing generalization. The CLIP-VV visual encoder captures local visual details with V-V attention, while Conv-Adapter and Bi-CMCI provide comprehensive cross-modal fusion of global and local features. The red dashed line represents the $\mathcal{L}_{\text{KD}}$ loss function introduced by KnPL, guiding LNPs and LAPs learning during training.}
    \label{fig:KAnoCLIP_Overview}
    \end{figure}

\section{The Method}
\subsection{Overview}


     As shown in Figure \ref{fig:KAnoCLIP_Overview}, our KAnoCLIP framework introduces Knowledge-Driven Prompt Learning to remove the need for fixed text prompts and enhance generalization. By integrating CLIP-VV, Bi-CMCI, and Conv-Adapter, KAnoCLIP enhances local visual features, cross-modal interactions. These innovations significantly improve zero-shot anomaly detection performance.

     During training, the KAnoCLIP framework minimizes the loss function \(\mathcal{L}_{\text{total}}\) in Equation \ref{eq:loss_total} using an auxiliary anomaly detection dataset. During inference, a test image \( x_i \) is processed through the CLIP-VV visual encoder to extract patch features \( F^{i}_{\text{patch}} \). LNPs and LAPs, previously learned normal and abnormal text prompts, are inputted into the CLIP text encoder to extract text features $F_{\text{text}}$. These patch and text features are combined in the Bi-CMCI module to generate an anomaly map \( M_i \in \mathbb{R}^{H \times W} \) for each layer. The maps are summed and normalized to create the final anomaly map \( M \). After adjusting global visual features using the Conv-Adapter, the maximum value of \( M \) determines the global anomaly score.

    \subsection{Knowledge-Driven Prompt learning}

    CoOp-based prompt learning \cite{zhou2022learning} uses the pre-trained CLIP model for downstream tasks, but overfits to base class objects, reducing performance on unseen classes. KgCoOp \cite{yao2023visual} found that performance drops are linked to the distance between learnable (CoOp) and fixed (CLIP) prompts. By reducing this distance, generalization improves. Inspired by KgCoOp, we initially used fixed CLIP prompts (e.g., "a photo of an abnormal [class]") to guide learnable anomaly prompts, but this failed due to insufficient anomaly knowledge. We propose combining LLMs' general knowledge with VLMs' detailed image descriptions to create a knowledge-driven prompt learning (KnPL) method for zero-shot anomaly detection.
    
    \subsubsection{Constructing the LLM-VQA Knowledge Base}
    
    \begin{figure}[t!]
    \centering
    \includegraphics[width=0.5\linewidth]{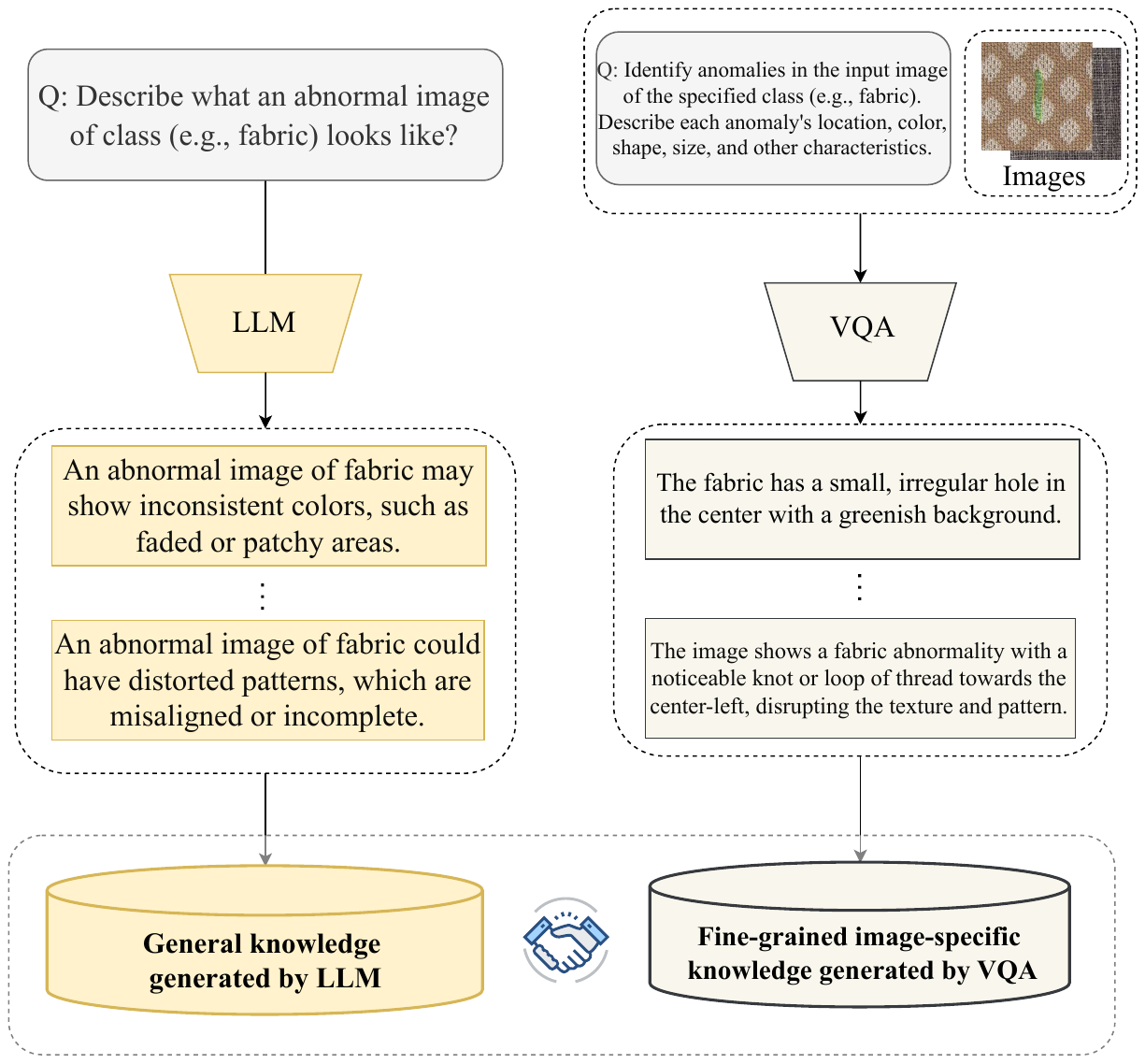}
    \caption{Constructing the LLM-VQA Knowledge Base: Generating Potential Anomalies and Image-Specific Descriptions.}
    \label{fig:KnPL}
    \end{figure}
    
    As shown in Figure \ref{fig:KnPL}, for each anomaly category, the LLM (GPT-3.5) generates $n$ optimal anomaly descriptions and background information (default $n$=5). The prompt template is: "Q: Describe what an abnormal image of {class (e.g., fabric)} looks like?" LLM-generated descriptions such as: "An abnormal image of fabric may show inconsistent colors, such as faded or patchy areas". The VQA model (Llama3)  generates $m$ image-specific anomaly descriptions per image (default $m$=1). The prompt template is: "<IMAGE> + Q: Identify anomalies in the input image of the specified {class (e.g., fabric)}. Describe each anomaly's location, color, shape, size, and other characteristics." VQA-generated descriptions such as: "The fabric has a small, irregular hole in the center with a greenish background". We refer to the anomaly descriptions generated by the LLM and VQA as the LLM-VQA knowledge base.

    
    \subsubsection{{Guiding Learnable Normal/Abnormal Prompts}}
    
    The knowledge-driven Learnable Normal Prompts (LNPs) and Learnable Abnormal Prompts (LAPs) are defined as follows:
        \begin{equation}
        \begin{aligned}
        P_n &= [N_1][N_2]\dots[N_K] [Normal][class]  \\
        P_a &= [A_1][A_2]\dots[A_K] [Abnormal][class] ,
         \end{aligned}
        \label{eq:NP}
    \end{equation} 
    where $N_i$ and $A_i$ ($i \in \{1, \dots, K\}$) are learnable word embeddings for normal and abnormal text prompts, respectively. $K$ denotes the length of the learnable prefix. [Normal] includes adjectives like "normal" and "perfect", while [Abnormal] includes "abnormal" and "defective". [class] represents the anomaly category to be detected.

     We introduce a Knowledge-Driven (KD) loss function, \(\mathcal{L}_{\text{KD}}\), to guide and constrain the generation of LNPs and LAPs. This loss minimizes the Euclidean distance between abnormal prompts generated by LLM-VQA and the LAPs, while simultaneously maximizing the distance to the LNPs.
    \begin{equation}
	\begin{aligned}
        \mathcal{L}_{KD}=
        \max \left(0,d(\frac{\bar{{w}}^{k}}{\|\bar{{w}}^{k}\|_2}, \frac{\bar{{w}}^{a}}{\|\bar{{w}}^{a}\|_2})-d(\frac{\bar{{w}}^{k}}{\|\bar{{w}}^{k}\|_2}, \frac{\bar{{w}}^{n}}{\|\bar{{w}}^{n}\|_2} )\right),
	\end{aligned}
	\label{tip}
    \end{equation}
    where $d(\cdot, \cdot)$ represent the Euclidean distance. $\bar{\textbf{w}}^{n}$ and $\bar{\textbf{w}}^{a}$ are the means of all LNPs/LAPs features, while $\bar{\textbf{w}}^{k}$ denotes the mean of the abnormal prompts features generated by LLM-VQA:
    \begin{equation}
	\begin{aligned}
        \bar{\textbf{{w}}}^k=\frac{\sum_{i=1}^{N} g(\textbf{p}^{LLM}_i)+\sum_{j=1}^{M}
        g(\textbf{p}^{VQA}_j)}{N + M} ,
	\end{aligned}
	\label{tip}
    \end{equation}
   where \(g(\cdot)\) be the CLIP text encoder. \(\textbf{p}^{LLM}_i\) and \(\textbf{p}^{VQA}_i\) represent abnormal text prompts generated by LLM and VQA, respectively, with \(N\) and \(M\) being their respective counts.



\subsection{Refining Local Visual Space with V-V Attention}
    
The original CLIP visual encoder, pre-trained with contrastive loss, produces global embeddings and uses a Q-K self-attention mechanism that disrupts local visual semantics, impairing fine-grained anomaly detection. To address this, we introduce a V-V attention mechanism \cite{li2023clip} into the CLIP visual encoder, enhancing the extraction of local visual features crucial for pixel-level anomaly segmentation without altering the original structure. V-V attention preserves local visual semantics by focusing on relationships between local features, resulting in attention maps with a distinct diagonal pattern. This enhanced visual encoder is termed CLIP-VV, and its feature extraction process is described as follows:
\begin{equation}
    \begin{aligned}
        Attention(\textbf{Q},\textbf{K},\textbf{V})=softmax(\textbf{Q} \cdot \textbf{K}^\text{T} \cdot scale) \cdot \textbf{V}
    \end{aligned}
\label{eq:qkv}
\end{equation}
\begin{equation}
    \begin{aligned}
    S_{l-1}^{ori} = [s_{cls}; s_1; s_2; \ldots; s_T]
    \end{aligned}
\end{equation}
\begin{equation}
\begin{aligned}
    S_{l-1} = [s_{cls}'; s_1'; s_2'; \ldots; s_T']
\end{aligned}
\end{equation}
\begin{equation}
    \begin{aligned}
    [Q_l, K_l, V_l] = [W_q S_{l-1}^{ori}, W_k S_{l-1}^{ori}, W_v S_{l-1}^{ori}].
    \end{aligned}
\end{equation}

using the V-V attention mechanism, the output of layer \( l \) is:
\begin{equation}
    \begin{aligned}
    S_l = Project_l(Attention(V_l, V_l, V_l)) + S_{l-1},
    \end{aligned}
\end{equation}
where \( S_{l-1}^{ori} \) is the original output and \( S_{l-1} \) is the local-aware output. \( Project_l \) are linear projections. Final outputs are \( S_l \). For anomaly detection, \( S_l{[0]} \) is used for image-level detection, and \( S_l{[1:]} \) for pixel-level detection.

\subsection{Enhancing Cross-Modal Feature Integration}


 \begin{figure}[t!]
        \centering
        \begin{minipage}[t]{0.48\linewidth}
            \centering
            \includegraphics[width=0.8\linewidth]{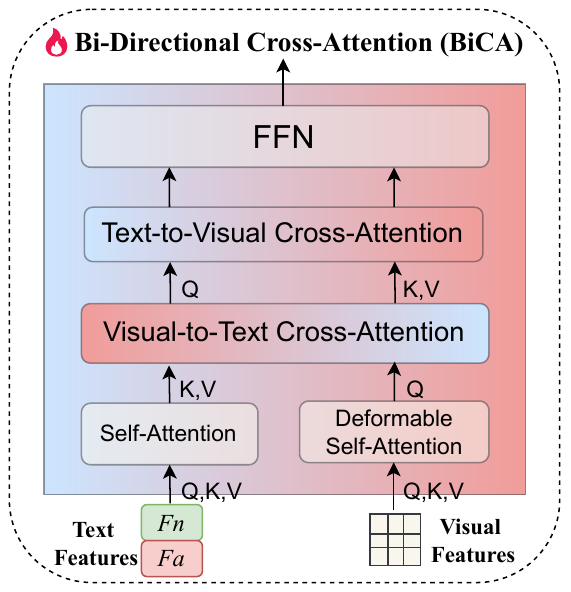}
            \caption{BiCA.}
            \label{fig:bi-cmci}
        \end{minipage}
        \hfill
        \begin{minipage}[t]{0.48\linewidth}
            \centering
            \includegraphics[width=0.6\linewidth]{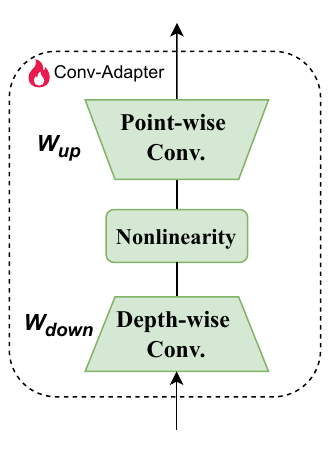}
            \caption{Conv-Adapter.}
            \label{fig:adapter}
        \end{minipage}
    \end{figure}

    We propose the Bi-Directional Cross-Attention for Multi-Level Cross-Modal Interaction (Bi-CMCI) and Conv-Adapter to address the limitations of traditional cross-modal feature fusion methods, such as linear layer projections \cite{chen2023zero}. These traditional methods often struggle to capture complex interactions and dependencies between visual and textual modalities, which are crucial for effective anomaly detection. Bi-CMCI manages local cross-modal interactions, while Conv-Adapter aligns global features. This dual approach effectively captures both local details and global context, thereby enhancing the model's accuracy in detecting subtle anomalies.

    \vspace{0.1em}
        
     \noindent \textbf{Bi-CMCI.} As illustrated in Figure \ref{fig:KAnoCLIP_Overview} and \ref{fig:bi-cmci}, Bi-CMCI employs a bi-directional cross-attention (BiCA) \cite{hou2019cross} mechanism to align textual descriptions with visual parts through Text-to-Visual Cross-Attention and refine text with local visual features via Visual-to-Text Cross-Attention. Self-Attention ensures independent understanding within each modality, while Deformable Self-Attention \cite{xia2022vision} adapts to key visual parts, addressing variations in shape and position. Bi-CMCI operates across the four stages of the CLIP-VV visual encoder to capture multi-level detailed information for identifying subtle anomalies. Each stage extracts intermediate patch-level features \(F^{i}_{patch} \in \mathbb{R}^{H_i W_i \times C_i}\). The LNPs and LAPs are input into the CLIP Text Encoder \(g(\cdot)\), producing text features representing normal and abnormal cases, \(F_{text} = [F_{n}, F_{a}] \in \mathbb{R}^{2 \times C}\). The Bi-CMCI module then deeply fuses the cross-modal features \(F^{i}_{patch}\) and \(F_{text}\), resulting in the normal map \(M^n_i \in \mathbb{R}^{H \times W}\), anomaly map \(M^a_i \in \mathbb{R}^{H \times W}\), and the final localization result \(M\).

    \begin{equation}
    \begin{aligned}
    M^n, M^a = Norm (\sum_{i=1}^{4} {BiCA_{i}(F^{i}_{\text{patch}}, F_{\text{text}}}))
    \end{aligned}
    \label{eq:localization result}
    \end{equation}
    \begin{equation}
    M = \frac{G_\sigma (M^a + 1 - M^n)}{2},
    \label{eq:final_m}
    \end{equation}
    \noindent where \( \operatorname{BiCA_{i}}(\cdot, \cdot) \) represents the Bi-Directional Cross-Attention operation between the local visual and textual features at the \(i\)-th stage. \( \operatorname{Norm}(\cdot) \) normalizes the anomaly map values to the range of 0 to 1. \( G_\sigma \) is a Gaussian filter with \( \sigma \) controlling the smoothing.
    
     \vspace{0.1em}
 
     \noindent \textbf{Conv-Adapter.} As illustrated in Figure \ref{fig:adapter}, the Conv-Adapter uses a depthwise separable convolution bottleneck \cite{kaiser2017depthwise} to better align global visual and textual features. It includes three main components: depthwise convolution for capturing spatial features, activation functions for nonlinearity, and pointwise convolution for dimensionality reduction and efficiency. The Conv-Adapter and global anomaly score is calculated as:
    \begin{equation}
    \begin{aligned}
    \hat{I}_{G} = \text{ReLU}(\text{LN}(I_{G}) W_{\text{down}}) W_{\text{up}}
    \end{aligned}
    \end{equation}
    \begin{equation}
        \begin{aligned}
        S_{global} = \text{softmax}(\hat{I}_{G} \cdot F_{text}^T) + \max(M),
        \end{aligned}
    \label{S_global}
    \end{equation}
    \noindent where \(I_{G}\) denotes the global visual feature, \(\text{LN}\) represents layer normalization, \(W_{\text{down}} \in \mathbb{R}^{d \times d_{\text{bottle}}}\) is the down-projection weight matrix, and \(W_{\text{up}} \in \mathbb{R}^{d_{\text{bottle}} \times d}\) is the up-projection weight matrix. The ReLU activation function is denoted as \(\text{ReLU}\). The output feature after the Conv-Adapter transformation is \(\hat{I}_{G}\). \(M\) is the anomaly map calculated in equation \ref{eq:final_m}, and \(\max(\cdot)\) denotes the maximum operation.

\subsection{Joint Optimization}

We develop joint optimization, a method for effectively learning knowledge-driven text prompts from both global and local perspectives. The total loss function \( \mathcal{L}_{\text{total}} \) is defined as follows:
\begin{equation}
\begin{aligned}
\mathcal{L}_{\text{total}} = \alpha \mathcal{L}_{\text{KD}} + \beta \mathcal{L}_{\text{global}} + \gamma \mathcal{L}_{\text{local}},
\end{aligned}
\label{eq:loss_total}
\end{equation}
where the hyperparameters \( \alpha \), \( \beta \), and \( \gamma \) balance the three loss components, and we default all of them to 1.

\noindent\textbf{Global Loss (\( \mathcal{L}_{\text{global}} \))}.
   The global loss is a binary cross-entropy loss that matches the cosine similarity between text embeddings and visual embeddings from auxiliary data of normal/anomalous images.
\begin{equation}
\begin{aligned}
    \mathcal{L}_{\text{global}} = \operatorname{BCE}(S_{\text{global}}, \text{\textit{Label}}),
\end{aligned}
\end{equation}

\noindent where \( S_{global} \) represents the global anomaly score as calculated in equation \ref{S_global}, and \( Label \) indicates whether the image is anomalous.

\noindent\textbf{Local Loss (\( \mathcal{L}_{\text{local}} \))}.
   The local loss combines focal loss \cite{lin2017focal} and dice loss \cite{li2019dice} to handle fine-grained local anomalous regions in the intermediate layers of the visual encoder.
\begin{equation}
    \begin{aligned}
    \mathcal{L}_{\text{local}} = \operatorname{Focal}(M^{a}, G) + \operatorname{Dice}(M^{n}, I-G) + \operatorname{Dice}(M^{a}, G),
    \end{aligned}
\end{equation}
\noindent where \( G \in \mathbb{R}^{H \times W} \) is the ground truth segmentation mask, with \( G_{ij} = 1 \) for anomalous pixels and \( G_{ij} = 0 \) otherwise. \( I \) denotes a matrix of ones.

\begin{table*}[]
\caption{ZSAD Performance comparison in industrial domain. The best-performing result is highlighted in bold and red, while the second-best is highlighted in blue.}
\label{tab:industry_results}
\centering
\renewcommand{\arraystretch}{0.9} 
\resizebox{\linewidth}{!}{
\begin{tabular}{clclcccccccccccc}
\hline
\multicolumn{2}{c}{} & \multicolumn{2}{c}{} & \multicolumn{2}{c}{MVTec-AD} & \multicolumn{2}{c}{VisA} & \multicolumn{2}{c}{MPDD} & \multicolumn{2}{c}{BTAD} & \multicolumn{2}{c}{SDD} & \multicolumn{2}{c}{DAGM} \\ \cline{5-16} 
\multicolumn{2}{c}{\multirow{-2}{*}{Method}} & \multicolumn{2}{c}{\multirow{-2}{*}{Public}} & Image-AUC & Pixel-AUC & Image-AUC & Pixel-AUC & Image-AUC & Pixel-AUC & Image-AUC & Pixel-AUC & Image-AUC & Pixel-AUC & Image-AUC & Pixel-AUC \\ \hline
\multicolumn{2}{c}{CLIP} & \multicolumn{2}{c}{ICML 2021} & 74.1 & 38.4 & 66.4 & 46.6 & 54.3 & 62.1 & 34.5 & 30.6 & 65.7 & 39 & 79.6 & 28.2 \\
\multicolumn{2}{c}{CLIP-AC} & \multicolumn{2}{c}{IMCL 2021} & 71.5 & 38.2 & 65 & 47.8 & 56.2 & 58.7 & 51 & 32.8 & 65.2 & 32.5 & 82.5 & 32.7 \\
\multicolumn{2}{c}{WinCLIP} & \multicolumn{2}{c}{CVPR 2023} & {\color[HTML]{3166FF} 91.8} & 85.1 & 78.1 & 79.6 & 63.6 & 76.4 & 68.2 & 72.7 & 84.3 & 68.8 & 91.8 & 87.6 \\
\multicolumn{2}{c}{April-GAN} & \multicolumn{2}{c}{CVPR 2023} & 86.1 & 87.6 & 78 & 94.2 & 73 & 94.1 & 73.6 & 60.8 & 79.8 & 79.8 & 94.4 & 92.4 \\
\multicolumn{2}{c}{AnomalyCLIP} & \multicolumn{2}{c}{ICLR 2024} & 91.5 & {\color[HTML]{3166FF} 91.1} & 82.1 & {\color[HTML]{3166FF} 95.5} & {\color[HTML]{3166FF} 77} & 96.5 & {\color[HTML]{3166FF} 88.3} & {\color[HTML]{3166FF} 94.2} & {\color[HTML]{3166FF} 84.7} & 90.6 & {\color[HTML]{FE0000} \textbf{97.5}} & {\color[HTML]{3166FF} 95.6} \\
\multicolumn{2}{c}{MVFA-AD} & \multicolumn{2}{c}{CVPR 2024} & 88.9 & 89.3 & {\color[HTML]{3166FF} 82.3} & 94.8 & 76.4 & {\color[HTML]{3166FF} 96.6} & 80.1 & 81.4 & 82.5 & {\color[HTML]{3166FF} 91.7} & 95.8 & 95.1 \\ \hline
\multicolumn{2}{c}{Ours (KAnoCLIP)} & \multicolumn{2}{c}{—} & {\color[HTML]{FE0000} \textbf{93.1}} & {\color[HTML]{FE0000} \textbf{94.3}} & {\color[HTML]{FE0000} \textbf{83.8}} & {\color[HTML]{FE0000} \textbf{97.7}} & {\color[HTML]{FE0000} \textbf{77.8}} & {\color[HTML]{FE0000} \textbf{98.3}} & {\color[HTML]{FE0000} \textbf{90.6}} & {\color[HTML]{FE0000} \textbf{96.5}} & {\color[HTML]{FE0000} \textbf{86.2}} & {\color[HTML]{FE0000} \textbf{93.2}} & {\color[HTML]{3166FF} 97.3} & {\color[HTML]{FE0000} \textbf{96.8}} \\ \hline
\end{tabular}
}
\end{table*}

\begin{table*}[h]
\caption{ZSAD Performance comparison in medical domain. Since the Br35H, COVID-19, and HeadCT datasets lack pixel-level anomaly segmentation ground truth, we only performed anomaly classification on these three datasets.}
\label{tab:medical_results}
\centering
\renewcommand{\arraystretch}{0.9} 
\resizebox{\linewidth}{!}{
\begin{tabular}{clclcccccccccccc}
\hline
\multicolumn{2}{c}{} & \multicolumn{2}{c}{} & \multicolumn{2}{c}{BrainMRI} & \multicolumn{2}{c}{LiverCT} & \multicolumn{2}{c}{RESC} & \multicolumn{2}{c}{Br35H} & \multicolumn{2}{c}{COVID-19} & \multicolumn{2}{c}{HeadCT} \\ \cline{5-16} 
\multicolumn{2}{c}{\multirow{-2}{*}{Method}} & \multicolumn{2}{c}{\multirow{-2}{*}{Public}} & Image-AUC & Pixel-AUC & Image-AUC & Pixel-AUC & Image-AUC & Pixel-AUC & Image-AUC & Pixel-AUC & Image-AUC & Pixel-AUC & Image-AUC & Pixel-AUC \\ \hline
\multicolumn{2}{c}{CLIP} & \multicolumn{2}{c}{ICML 2021} & {\color[HTML]{333333} 53.9} & {\color[HTML]{333333} 71.7} & {\color[HTML]{333333} 57.1} & {\color[HTML]{333333} 82.3} & {\color[HTML]{333333} 38.5} & {\color[HTML]{333333} 70.8} & {\color[HTML]{333333} 78.4} & {\color[HTML]{333333} -} & {\color[HTML]{333333} 73.7} & {\color[HTML]{333333} -} & {\color[HTML]{333333} 56.5} & - \\
\multicolumn{2}{c}{CLIP-AC} & \multicolumn{2}{c}{IMCL 2021} & {\color[HTML]{333333} 60.6} & {\color[HTML]{333333} 66.4} & {\color[HTML]{333333} 61.9} & {\color[HTML]{333333} 88.4} & {\color[HTML]{333333} 40.3} & {\color[HTML]{333333} 75.9} & {\color[HTML]{333333} 82.7} & {\color[HTML]{333333} -} & {\color[HTML]{333333} 75} & {\color[HTML]{333333} -} & {\color[HTML]{333333} 60.0} & - \\
\multicolumn{2}{c}{WinCLIP} & \multicolumn{2}{c}{CVPR 2023} & {\color[HTML]{333333} 66.5} & {\color[HTML]{333333} 86.0} & {\color[HTML]{333333} 64.2} & {\color[HTML]{333333} 92.2} & {\color[HTML]{333333} 42.5} & {\color[HTML]{333333} 80.6} & {\color[HTML]{333333} 80.5} & {\color[HTML]{333333} -} & {\color[HTML]{333333} 66.4} & {\color[HTML]{333333} -} & {\color[HTML]{333333} 81.8} & - \\
\multicolumn{2}{c}{April-GAN} & \multicolumn{2}{c}{CVPR 2023} & {\color[HTML]{333333} 76.4} & {\color[HTML]{333333} 91.8} & {\color[HTML]{333333} 70.6} & {\color[HTML]{333333} 94.1} & {\color[HTML]{333333} 75.6} & {\color[HTML]{333333} 85.2} & {\color[HTML]{333333} 93.1} & {\color[HTML]{333333} -} & {\color[HTML]{333333} 15.5} & {\color[HTML]{333333} -} & {\color[HTML]{333333} 89.1} & - \\
\multicolumn{2}{c}{AnomalyCLIP} & \multicolumn{2}{c}{ICLR 2024} & {\color[HTML]{333333} 78.3} & {\color[HTML]{3531FF} 92.2} & {\color[HTML]{333333} 73.7} & {\color[HTML]{3531FF} 96.2} & {\color[HTML]{333333} 83.3} & {\color[HTML]{333333} 91.5} & {\color[HTML]{FE0000} \textbf{94.6}} & {\color[HTML]{333333} -} & {\color[HTML]{333333} 80.1} & {\color[HTML]{333333} -} & {\color[HTML]{3166FF} 93.4} & {\color[HTML]{333333} -} \\
\multicolumn{2}{c}{MVFA-AD} & \multicolumn{2}{c}{CVPR 2024} & {\color[HTML]{3531FF} 78.6} & {\color[HTML]{333333} 90.3} & {\color[HTML]{3531FF} 74.2} & {\color[HTML]{333333} 95.9} & {\color[HTML]{3166FF} 84.2} & {\color[HTML]{3531FF} 92.0} & {\color[HTML]{333333} 90.5} & {\color[HTML]{333333} -} & {\color[HTML]{3166FF} 83.5} & {\color[HTML]{333333} -} & {\color[HTML]{333333} 91.3} & {\color[HTML]{333333} -} \\ \hline
\multicolumn{2}{c}{Ours (KAnoCLIP)} & \multicolumn{2}{c}{—} & {\color[HTML]{FE0000} \textbf{80.7}} & {\color[HTML]{FE0000} \textbf{93.5}} & {\color[HTML]{FE0000} \textbf{78.2}} & {\color[HTML]{FE0000} \textbf{98.6}} & {\color[HTML]{FE0000} \textbf{84.8}} & {\color[HTML]{FE0000} \textbf{93.5}} & {\color[HTML]{3531FF} 94.2} & {\color[HTML]{333333} \textbf{-}} & {\color[HTML]{FE0000} \textbf{85.4}} & {\color[HTML]{333333} \textbf{-}} & {\color[HTML]{FE0000} \textbf{95.8}} & {\color[HTML]{333333} \textbf{-}} \\ \hline
\end{tabular}
}
\end{table*}
\section{Experiments}

\subsection{EXPERIMENT SETUP}

\textbf{Datasets.} We conducted experiments on 12 datasets, covering industrial and medical anomaly detection. Industrial datasets included MVTec-AD \cite{bergmann2019mvtec} , VisA \cite{zou2022spot}, MPDD \cite{jezek2021mpdd}, BTAD \cite{mishra2021vt}, SDD \cite{tabernik2020segmentation}, and DAGM \cite{wieler2007weakly}. Medical datasets were BrainMRI \cite{salehi2021multiresolution}, LiverCT \cite{bilic2023liver}, RESC \cite{hu2019automated}, Br35H \cite{bourennane2023deep}, COVID-19 \cite{rahman2021exploring}, and HeadCT \cite{salehi2021multiresolution}.  

\noindent \textbf{Evaluation Metrics.} We used the Area Under the Receiver Operating Characteristic Curve (AUC) for evaluation, applying image-level AUC for classification and pixel-level AUC for segmentation.

\noindent \textbf{Baselines.} Our KAnoCLIP was thoroughly compared with state-of-the-art methods, including CLIP \cite{radford2021learning}, CLIP-AC \cite{radford2021learning}, WinCLIP \cite{jeong2023winclip}, APRIL-GAN \cite{chen2023zero}, MVFA-AD \cite{huang2024adapting}, and AnomalyCLIP \cite{zhou2023anomalyclip}. 

\noindent \textbf{Implementation details.}  We used the CLIP model (VIT-L/14@336px) as our backbone, keeping all parameters frozen. Following APRIL-GAN and AnomalyCLIP, we trained our model on the MVTec-AD test data and evaluated the ZSAD performance on other datasets. Learnable word embeddings length was set to 12. Experiments were conducted in PyTorch 2.0.0 on a single NVIDIA RTX 3090 GPU.

\subsection{MAIN RESULTS}
  

\subsubsection{ZSAD Performance Comparison}

\noindent As illustrated in Tables \ref{tab:industry_results} and \ref{tab:medical_results}, our KAnoCLIP model has achieved state-of-the-art results across 12 industrial and medical anomaly detection datasets, surpassing the performance of MVFA-AD and AnomalyCLIP. Notably, in the prominent MVTec-AD and VisA datasets, our method improved pixel-level AUC scores by \textbf{3.2} and \textbf{2.2}, respectively, and enhanced image-level AUC scores by \textbf{1.3} and \textbf{1.5}. This highlights our superior performance in both anomaly segmentation and anomaly classification. Additionally, across six medical datasets, KAnoCLIP exhibited significant improvements, particularly in BrainMRI and LiverCT, with image-level AUC increases of \textbf{2.1} and \textbf{4.0} and pixel-level AUC enhancements of \textbf{1.3} and \textbf{2.4}. Overall, KAnoCLIP has demonstrated robust effectiveness and generalization across a diverse range of anomaly detection datasets.



\subsubsection{Results Analysis and Discussion}
The original CLIP model performed poorly because its text prompts were designed for image classification, not anomaly detection. CLIP-AC made slight improvements by adding "normal" and "abnormal" descriptors but still fell short. WinCLIP and APRIL-GAN, with manual crafted anomaly detection prompts, performed better. AnomalyCLIP, using learnable anomaly text prompts, further improved zero-shot performance but overfit to base classes without general knowledge guidance. MVFA-AD, an adapter for WinCLIP, enhanced cross-domain performance but lacked learnable prompts and effective cross-modal interactions. Our KAnoCLIP method achieved the best results by improving cross-modal interactions and integrating global-local context, making it highly effective for zero-shot anomaly detection in industrial and medical applications.
\subsection{ABLATION STUDY}
\label{sec:ABLATION STUDY}

\begin{table}[]
\caption{The ablation results for our KAnoCLIP framework demonstrate the incremental effects of adding each module.}
\label{tab:main_ablation_results}
\centering
\renewcommand{\arraystretch}{0.8}
\resizebox{\linewidth}{!}{
\begin{tabular}{@{}cccccccc@{}}
\toprule
 &  &  &  & \multicolumn{2}{c}{MVTec-AD} & \multicolumn{2}{c}{VisA} \\ \cmidrule(l){5-8} 
\multirow{-2}{*}{KnPL} & \multirow{-2}{*}{CLIP-VV} & \multirow{-2}{*}{Bi-CMCI} & \multirow{-2}{*}{Conv-Adapter} & Image-AUC & Pixel-AUC & Image-AUC & Pixel-AUC \\ \midrule
\XSolidBrush & \XSolidBrush & \XSolidBrush & \XSolidBrush & 66.8 & 55.4 & 58.2 & 47.1 \\
\Checkmark & \XSolidBrush & \XSolidBrush & \XSolidBrush & 85.9 & 87.4 & 78.6 & 91.2 \\
\Checkmark & \Checkmark & \XSolidBrush & \XSolidBrush & 87.5 & 88.9 & 80.3 & 92.3 \\
\Checkmark & \Checkmark & \Checkmark & \XSolidBrush & 92.0 & 92.4 & 82.3 & 95.8 \\
\Checkmark & \Checkmark & \Checkmark & \Checkmark & {\color[HTML]{FE0000} \textbf{93.1}} & {\color[HTML]{FE0000} \textbf{94.3}} & {\color[HTML]{FE0000} \textbf{83.8}} & {\color[HTML]{FE0000} \textbf{97.7}} \\ \bottomrule
\end{tabular}
}
\end{table}

We conducted ablation experiments on the MVTec-AD and VisA datasets to validate the effectiveness of the four key modules in the KAnoCLIP framework: KnPL, CLIP-VV, Bi-CMCI, and Conv-Adapter. Each module significantly enhanced zero-shot anomaly detection, with KnPL providing the most substantial improvement. Detailed results are in Table \ref{tab:main_ablation_results}.



\section{Conclusion} 

In this paper, we introduce KAnoCLIP, a zero-shot anomaly detection framework using KnPL, which integrates LLM's general knowledge with VQA system's image-specific insights to create learnable anomaly prompts, eliminating fixed text prompts and enhancing generalization to new anomaly classes. KAnoCLIP employs three modules: CLIP-VV, Bi-CMCI, and Conv-Adapter, optimizing visual space, cross-modal interactions, and global-local context integration. Extensive experiments on 12 industrial and medical datasets show that KAnoCLIP consistently outperforms SOTA methods, demonstrating superior generalization capabilities.

\bibliographystyle{IEEEbib}
\bibliography{mybibliography}

\end{document}